# Temporal-Difference Networks for Dynamical Systems with Continuous Observations and Actions

**Christopher M. Vigorito**
Computer Science
University of Massachusetts
Amherst, MA
vigorito@cs.umass.edu

## Abstract

Temporal-difference (TD) networks are a class of predictive state representations that use well-established TD methods to learn models of partially observable dynamical systems. Previous research with TD networks has dealt only with dynamical systems with finite sets of observations and actions. We present an algorithm for learning TD network representations of dynamical systems with continuous observations and actions. Our results show that the algorithm is capable of learning accurate and robust models of several noisy continuous dynamical systems. The algorithm presented here is the first fully incremental method for learning a predictive representation of a continuous dynamical system.

## 1 Introduction

Predictive representations of state are a class of generative models that represent a dynamical system in terms of a set of predictions about sequences of observations generated by that system (Littman et al., 2002). Recent work has shown that certain formalizations of predictive representations are strictly more expressive than other models of discrete dynamical systems that use historical information or probabilistic distributions over unobservable variables as a representation (e.g., k-Markov models, POMDPs) (Singh & James, 2004). Empirically it has also been shown that in certain domains predictive representations can lead to better generalization than other representations (Rafols et al., 2005). In addition to this theoretical and practical appeal, predictive representations have the desirable property of being *grounded* in the sense that the representation is defined exclusively in terms of observable quantities. Though they share this property with other representations, such as k-Markov models, they are more expressive than such models.

One formalism for predictive representations is the Temporal-difference (TD) network (Sutton & Tanner, 2005). TD networks use well-established TD learning methods to incrementally update the predictions that define their state based on a stream of successive observations. All previous research with TD networks has focused on modeling dynamical systems with discrete observations and actions. Although there has been some work on other formalisms of predictive representations in continuous systems (Wingate, 2008), these approaches have not yet been extended to a fully online, incremental setting.

The work presented here is intended to fill this gap, providing a method for incrementally learning predictive representations in continuous dynamical systems. We achieve this by adapting the TD network formalization to make predictions about the values of feature functions defined over the observation space of a dynamical system, and present a method for conditioning those predictions on actions that also take on continuous values. The ability to learn predictive representations of continuous dynamical systems online has potential applications in fields such as reinforcement learning, robotics, and time series prediction, where variables of interest may take on a continuum of values and data is received as a continuing stream of incoming observations.

In the following section we provide a formal definition of discrete and continuous dynamical systems. In section 3 we outline the TD network formalism and describe previous work with TD networks. Section 4 presents our modification to the TD network architecture that allows for continuous variables. We present results in noisy, continuous dynamical systems in section 5 and discuss our findings and future work in section 6.

## 2 Dynamical Systems

We present two formalizations of controlled, discrete-event dynamical systems (DEDS), one with discrete observations and actions, and one with continuous observations and actions. We do not consider continuous-time systems in this work. A DEDS with discrete observations and actions consists of two finite sets of symbols, $\mathcal{O}$ and $\mathcal{A}$, the observations and actions, respectively, and a dynamics function $P$, described below. At each discrete time step $t$ the system outputs an observation $o_t \in \mathcal{O}$ and accepts an action $a_t \in \mathcal{A}$. The alternating sequence of observations and actions that begins with the observation output at time 0 and ends with the action accepted at time $t$ is the history $h_t \in \mathcal{H}$ of the system at time $t$, where $\mathcal{H}$ is an infinite set of all possible histories. Each observation is chosen probabilistically according to $P : \mathcal{O} \times \mathcal{H} \to [0, 1]$, where $P(o, h)$ gives the probability that $o_{t+1} = o$ for some $o \in \mathcal{O}, h \in \mathcal{H}$. Thus, $P$ induces a probability mass function over $\mathcal{O}$, from which an observation is sampled at each time step.

A DEDS with continuous observations and actions is composed of two infinite sets, $\mathcal{O} \subseteq \Re^o$ and $\mathcal{A} \subseteq \Re^a$, again representing the observations and actions, where $o$ and $a$ are the dimensionalities of the observation and action spaces, respectively. As above, histories are alternating sequences of observations and actions, the only difference being that each observation and action is a point in its associated vector space rather than a discrete symbol. The dynamics function $P$ is defined in the same way, but now induces a probability density function over $\mathcal{O}$ from which each observation is sampled at every time step.

It should be noted that although the formalization of a discrete DEDS given above assumes a single observation is output by the system at each time step, it is trivial to extend this to vector-valued, discrete observations. Additionally, the formalization of an uncontrolled DEDS with either continuous or discrete osbervations is the same as above, but excludes the action set $\mathcal{A}$. Histories are then comprised of sequences of observations, and the definition of the dynamics function $P$ remains the same in each case. We next describe previous work with TD networks for modeling a discrete DEDS, and then present our method for modeling a continuous DEDS using a continuous TD network.

## 3 TD Networks

Being a predictive representation, a TD network maintains state by updating at each time step the probabilities of a set of predictions, or questions about the system the network models. The semantics of these predictions are realized by a question network, which defines the TD relationships between different predictions. The question network is a set of nodes, each representing a specific scalar prediction about some observation of the system some number of time steps in the future. Example question networks for an uncontrolled and controlled dynamical system are shown in Figure 1.

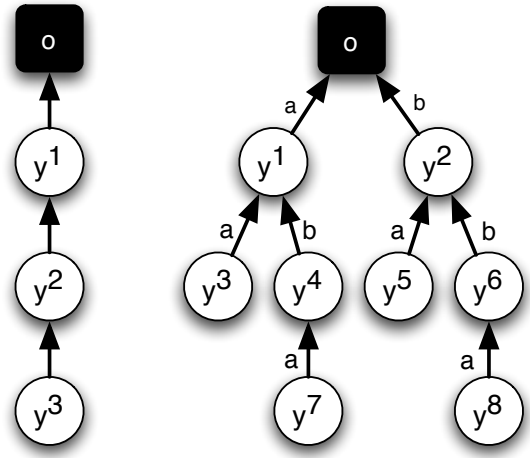

Figure 1: An example TD question network for an uncontrolled (left) and controlled (right) DEDS.

The links between nodes in the question network provide the target for each prediction—the quantity it attempts to predict. This quantity may be defined in terms of another prediction (circles in Figure 1), observation data (squares), or both. For example, node $y^1$ in the left network of Figure 1 might make a prediction about the probability that the observation will be some specific value (e.g., 1 if the observation is binary) at the next time step.

If we let $y_t^i$ denote the prediction of node $y^i$ at time $t$, and $z_t^i$ denote the target of $y_t^i$ at time $t$, then in this example $z_t^1 = Pr(o_{t+1} = 1)$. In contrast, node $y^2$ makes a prediction about the expected value of node $y^1$ at the next time step, and its target at time $t$ is thus $z_t^2 = E(y_{t+1}^1)$. Note that this has the same meaning as predicting the probability that the observation will be 1 at time $t + 2$.

For controlled systems, predictions can be conditioned on actions, as seen in the right-hand network in Figure 1. The system modeled by this network has two actions, $a$ and $b$, and action conditional predictions are indicated via labels on the links of the network. For example, node $y^1$ makes a prediction about the value of the observation at time $t + 1$ given that the agent takes action $a$ at time $t$. Similarly, node $y^4$ makes a prediction about the value of the observation at time $t+2$ given that the agent takes action $b$ and then action $a$ starting at time $t$. As in previous work, for ease of exposition we discuss only single-target question networks, i.e., nodes with only one parent.

The actual values of the predictions semantically defined by the question network are computed by a separate function approximator called the answer network. The state of the system, or output of the answer network is the vector of predictions $\mathbf{y}_t \in \Re^n$, where $n$ is the number of nodes in the network. At each time step an input vector $\mathbf{x}_t$ is computed as some function of the previous predictions $\mathbf{y}_{t-1}$, the previous action $a_{t-1}$, and the newly received observation $o_t$:

$$\mathbf{x}_t = \mathbf{x}(\mathbf{y}_{t-1}, a_{t-1}, o_t) \in \Re^m. \quad (1)$$

The prediction vector $\mathbf{y}_t$ is then computed as some function $\mathbf{u}$ of $\mathbf{x}_t$ and a modifiable parameter $\mathbf{W}$:

$$\mathbf{y}_t = \mathbf{u}(\mathbf{x}_t, \mathbf{W}) \in \Re^n. \quad (2)$$

A stochastic gradient descent update rule is used to modify the weights $w^{ij}$ of the network according to

$$\Delta w^{ij} = \alpha (z_t^i - y_t^i) c_t^i \frac{\partial y_t^i}{\partial w^{ij}}, \quad (3)$$

where $\alpha$ is a step size parameter, $c_t^i$ is an action condition defined below, and $z_t^i$ is the $i^{th}$ element of the target vector $\mathbf{z}_t$, which is computed as some function $\mathbf{z}$ of the latest observation and predictions:

$$\mathbf{z}_t = \mathbf{z}(o_t, \mathbf{y}_t) \in \Re^n. \quad (4)$$

The action conditions $\mathbf{c}_t$ determine the degree to which each prediction is responsible for matching its target given the agent's behavior, as defined by the question network. Formally, $\mathbf{c}_t$ is defined to be some function $\mathbf{c}$ of the previous action and predictions

$$\mathbf{c}_t = \mathbf{c}(a_{t-1}, \mathbf{y}_{t-1}) \in [0,1]^n. \quad (5)$$

Tanner and Sutton (2005) introduced TD($\lambda$) networks, which incorporate eligibility traces to deal with certain shortcomings of conventional TD networks. Eligibility traces were originally introduced in Sutton (1988) to provide a mechanism for making more general $n$-step backups of predictions in conventional TD learning, rather than the traditional 1-step backups. The parameter $\lambda \in [0,1]$ controls the degree to which longer sequences of predictions act as a target for learning.

When $\lambda = 0$, the target is simply the 1-step prediction. When $\lambda = 1$, the longest possible sequence of predictions is used as a target and given the full weight of each update. Intermediate values of $\lambda$ result in exponentially weighted averages of sequences of varying lengths being used as targets.

Notation for node targets in TD($\lambda$) networks makes use of the parent function $p(i)$, which denotes the parent of node $y^i$ as defined by a single link in the question network. Later parents of a node are denoted as $\{p(p(i)), p(p(p(i))), \ldots\}$, or $\{p^2(i), p^3(i), \ldots\}$ in short form, so that $p^k(i)$ identifies the $k^{th}$ parent of node $y^i$. The machinery necessary for incorporating eligibility traces is slightly more complicated for TD($\lambda$) networks than for the TD($\lambda$) algorithm used for value-function learning. A trace for each prediction $y_t^i$ must be maintained, and at each step the algorithm checks to see that the sequence of recently executed actions matches the conditions for the prediction, eliminating the trace if this is not the case.

Given a trace initialized at time step $t-k$ whose action conditions over the last $k$ time steps have been satisfied, the weights $\mathbf{W}$ are updated at time step $t$ using the temporal difference information $\mathbf{y}_t - \mathbf{y}_{t-1}$ and the past input vector $\mathbf{x}_{t-k}$ to improve the past prediction $\mathbf{y}_{t-k}$, with the update scaled appropriately by $\lambda^{t-k-1}$. We have only presented essential notation and intuition here, and refer the reader to Tanner and Sutton (2005) for the full details of the algorithm. In the following section we present our modified TD($\lambda$) algorithm, which allows for continuous observations and actions.

## 4 Continuous TD Networks

In the case of a dynamical system with continuous observations, one can no longer construct question networks to specify predictions of all possible values of the system's observations, since this would require infinitely many predictions. The solution we employ is to make predictions about the expected values of a set of feature functions defined over the observation space. More formally, we maintain a set $\Phi$ of feature functions $\phi_i : \Re^o \to \Re$, each element of which outputs at time $t$ a scalar value $\phi_i(o_t)$, where $o_t$ is the observation at time $t$ and $o$ is the dimensionality of the observation space. Predictions of the values of these functions, which together define state, can then be used as features for approximating other functions, e.g, value functions in a reinforcement learning setting (Sutton & Barto, 1998).

Figure 2 shows an illustration of a possible question network for an uncontrolled DEDS with continuous observations. The feature functions in this case can be thought of as radial basis functions with spherical co-variance matrices evenly tiled over a two-dimensional observation space. Each feature function acts as a target observation and each node predicts the expected value of one of the functions some number of time steps in the future. The semantics of the links are the same as those defined for discrete TD networks.

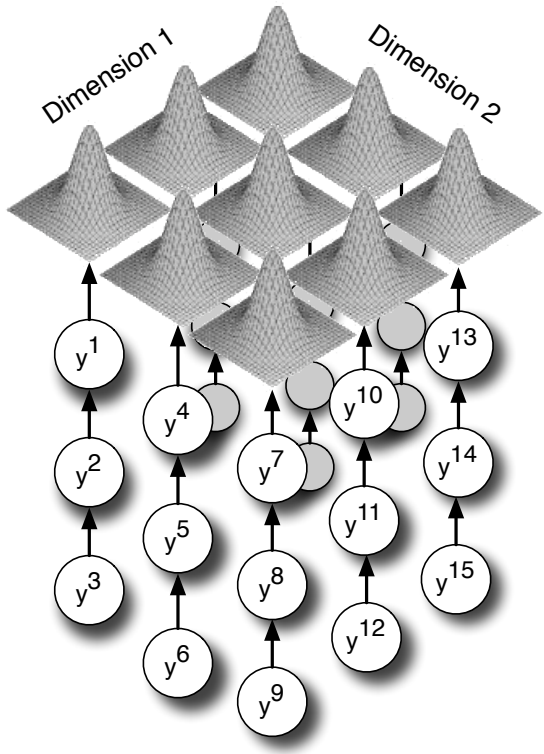

Figure 2: A example TD question network for an uncontrolled DEDS with continuous observations.

We must also provide a method for dealing with continuous actions in this framework, since it is not possible to have infinitely many action conditions in the question network. In all previous work with TD networks the action conditions were assumed to be binary since the action set was finite and small. The definition of the action conditions given in (5), however, is more general and allows for real-valued action conditions between 0 and 1.

To handle action conditions that allow for generalization over similar actions, we assume a set $\Psi$ of activation functions $\psi_i : \Re^a \to \Re$, each element of which takes an action in $\Re^a$ and provides a scalar value in $[0, 1]$ according to some similarity metric, indicating the degree to which that action matches the associated activation function. Each link of the question network may thus be conditioned on a particular activation function $\psi_i \in \Psi$ just as links are conditioned on

$\Phi \leftarrow$ set of observation feature functions
$\Psi \leftarrow$ set of action activation functions
$\mathbf{y} \leftarrow$ initial state vector
$\mathbf{W} \leftarrow$ initial weight matrix
$Traces \leftarrow \{\}$
**for** $t=0$ to $T$ **do**
  $newTraces \leftarrow \{\}$
  $\mathbf{a} \leftarrow chooseAction()$
  $\mathbf{o} \leftarrow getObservation(\mathbf{a})$
  $\mathbf{x}_t \leftarrow x(\mathbf{a}, \mathbf{o}, \mathbf{y}_{t-1}, \Phi, \Psi)$
  $\mathbf{y}_t \leftarrow \mathbf{W}\mathbf{x}_t$
  **for** $(i,k) \in Traces$ **do**
    **if** $p^{t-k}(i) \neq observation$ **then**
      $z \leftarrow \mathbf{y}_t[p^{t-k}(i)]$
    **else**
      $z \leftarrow \phi_{p^{t-k}(i)}(o)$
    **end if**
    $p \leftarrow \mathbf{y}_{t-1}[p^{t-k-1}(i)]$
    $c_k \leftarrow traceCondition(i,k) \cdot \psi_{p^{t-k-1(i)}}(\mathbf{a})$
    **for** $w^j \in W[i]$ **do**
      $w^j += \alpha(z-p)c_k x_k^j \lambda^{t-k-1}$
    **end for**
    **if** $p^{t-k}(i) \neq observation$ **then**
      $traceCondition(i,k) \leftarrow c_k$
      $newTraces \leftarrow newTraces \cup (i,k)$
    **end if**
  **end for**
  **for** $i \in \mathbf{y}$ **do**
    $traceCondition(i,t) \leftarrow 1$
    $newTraces \leftarrow newTraces \cup (i,t)$
  **end for**
  $Traces \leftarrow newTraces$
**end for**

Figure 3: Psuedo-code for the TD($\lambda$) network learning algorithm for continuous dynamical systems. Modified from Tanner and Sutton (2005).

specific actions in a discrete TD network. The value of the action condition for node $y^i$ at time $t$ is computed as $c_t^i = \psi_i(a_{t-1})$, where $\psi_i$ is the activation function on which node $y_i$ is conditioned.

We use Euclidean distance as a similarity metric in this work and employ radial basis functions as our activation functions for their ease of use. The action space is tiled evenly by these functions, with each $\psi_i$ having a different center $\mu_i \in \Re^a$ and $a \times a$ covariance matrix $\Sigma$. The value of each $\psi_i$ given action $a_t$ is thus computed as

$$\psi_i(a_t) = e^{-(\mathbf{a}_t - \mu_i)^T \Sigma_i^{-1}(\mathbf{a}_t - \mu_i)/2}. \qquad (6)$$

Although this allows for general covariance matrices, in our experiments we use spherical covariance matri-ces so that $\boldsymbol{\Sigma} = \sigma\mathbf{I}$, where $\mathbf{I}$ is the identity matrix and $\sigma$ is a parameter that determines the kernel width.

Figure 3 shows the TD($\lambda$) network learning algorithm for systems with continuous observations and actions. The algorithm is modified from Tanner and Sutton (2005). The first major difference is the construction of the answer network's input vector $\mathbf{x}_t$. Our approach constructs $\mathbf{x}_t$ by concatenating $\mathbf{y}_{t-1}$ with a vector containing the values of the observation basis functions $\phi_i \in \Phi$ and of the action activation functions $\psi_i \in \Psi$ at time $t$ given $\mathbf{o}_t$ and $\mathbf{a}_{t-1}$. The size of the input vector is thus $|\mathbf{y}| + |\Phi| + |\Psi|$. This is in contrast with previous work in which the input vector contained binary elements corresponding to each possible action-observation pair.

The answer network in previous work with TD networks was implemented as a generalized linear model, so that

$$\mathbf{y}_t = \sigma(\mathbf{W}\mathbf{x}_t) \in \Re^n, \qquad (7)$$

where the parameter $\mathbf{W}$ was a $|\mathbf{y}| \times |\mathbf{x}|$ weight matrix, and $\sigma$ was the vector-valued logistic function $\sigma(x) = \frac{1}{1+e^{-x}}$ applied element-wise to $\mathbf{W}\mathbf{x}_t$. In our work, however, nodes do not predict probabilities of binary predicates corresponding to discrete observations, and so the use of the logistic function to filter $\mathbf{W}\mathbf{x}_t$ is not appropriate. We thus let $\sigma$ be the identity function, resulting in a simple linear function approximator.

The final distinction concerns the method of updating eligibility traces. Because we use non-binary action conditions, traces are not eliminated as they are in the discrete algorithm when action conditions fail. Rather the action condition values must appropriately weight the updates associated with each trace based on the agent's recent actions. In order to achieve this, each trace must store an accumulated action condition that is initialized to 1 when the trace is created, and updated at each time step by multiplying it by the action condition value at the current time step. The function $traceCondition(\cdot)$ in the algorithm of Figure 3 represents the action condition value currently associated with a given trace.

## 5 Experiments

To evaluate our approach we tested our algorithm on a small set of partially observable, continuous dynamical systems. The systems are partially observable in the sense that the most recent observation does not provide enough information to maintain state; i.e., it is not a sufficient statistic for history. Multi-step predictions are thus needed to model the systems accurately.

For all of our experiments we employed radial basis functions (RBFs) for our observation features $\Phi$ as well as for our set of action activation functions $\Psi$. The action conditions were thus computed as given in (6), and the observation features, similarly, as

$$\phi_i(o_t) = e^{-(\mathbf{o}_t - \mu_i)^T \boldsymbol{\Sigma}_i^{-1}(\mathbf{o}_t - \mu_i)/2}, \qquad (8)$$

where $o_t$ is the observation at time $t$. The feature and activation function centers were tiled evenly over the observation space and action space, respectively, so that a system with observation dimension $o$ and action dimension $a$ had $n^o$ feature functions and $m^a$ activation functions, where $n$ and $m$ are the number of functions used per dimension for the observation and action spaces, respectively. As mentioned above we used spherical covariances for each of the functions so that $\boldsymbol{\Sigma} = \sigma\mathbf{I}$. We report the value of $n$, $m$, and $\sigma$ for each experiment as it is discussed.

Although one would ideally like to automate the construction of the question network when learning TD networks, to keep clear our focus on learning the parameters of a TD network for a continuous system we have left the issue of question network discovery in continuous TD networks to future work. The choice of question network for each system was thus made according to intuition and some trial and error based on knowledge of the systems being modeled.

Though we experimented with a few different types of question networks, we wound up using ones of the form shown in Figure 4 for each system. Each feature function $\phi_i \in \Phi$ was the parent (target) of $|\Psi|$ nodes, each of which, conditioned on a distinct $\psi_j \in \Psi$, predicted the value of $\phi_i$ one time step in the future. We did not implement a fully conditional network (which would require a number of nodes exponential in $|\Psi|$), but rather found that successive, identical action conditions chained off of each of the children of the observation nodes, as diagrammed in Figure 4, performed well. This structure is similar to previous question network structures used in some discrete TD networks (Tanner & Sutton, 2005). The trailing dots at the bottom of the figure indicate that the depth of each chain can vary. We used the same chain depth $d$ for each chain in a given question network, and report that depth for each experiment below. All experiments used a step size parameter $\alpha = 0.01$ and eligibility parameter $\lambda = 1$.

### 5.1 Uncontrolled Systems

We first tested the ability of a TD network to learn models of uncontrolled dynamical systems. Since there are no actions in these systems the question networks look as they would in Figure 4 if $|\Psi| = 1$, so that each basis function $\phi_i$ has just one chain of $d$ descendants.

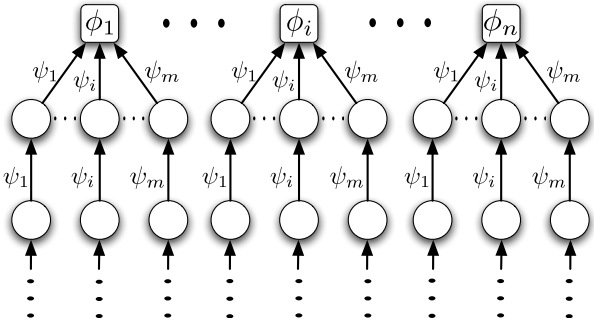

Figure 4: Form of the question networks used in our experiments. $|\Phi| = n$. $|\Psi| = m$.

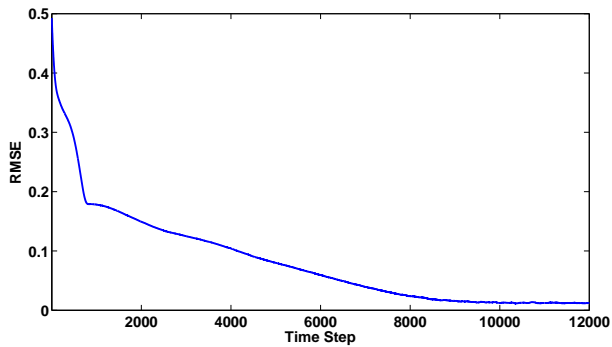

Figure 5: RMSE of the one-step predictions of all $\phi_i \in \Phi$ as a function of time for the noisy, uncontrolled square wave. Each point is an average of the error over the previous 100 time steps.

The first system was a simple square wave, which alternated between emitting one-dimensional values 0 and 1, each for five times steps at a time. Each observation emitted by the system was corrupted by mean-zero Gaussian noise with standard deviation 0.05. This is essentially a noisy, continuous analog of the cycle world presented in Tanner and Sutton (2005). We let $n = 4$, and $\sigma = 0.3$ for this experiment. In order to maintain state the network must have at least five steps of prediction, and so we used a depth $d = 5$ for each chain.

Figure 5 shows the average root-mean-squared error (RMSE) of the one-step predictions of the expected values of each feature function at each time step, averaged over all $|\Phi|$ predictions. The errors were computed by taking the difference between the predicted values of the basis functions at a given time step and the actual observed values of those functions at the next time step, squaring those errors, averaging them over all feature functions, and taking the square root of that average. Each point in the graph is an average of the RMSE for the previous 100 time steps.

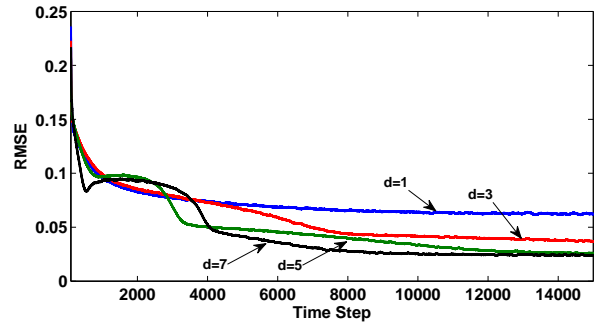

Figure 6: RMSE of the one-step predictions of all $\phi_i \in \Phi$ as a function of time for the noisy, uncontrolled sine wave. Each point is an average of the error over the previous 100 time steps.

The curve represents an average of 30 runs. We see that the network is able to learn a good model even given noisy observations with an amount of experience roughly equivalent to the amount taken to learn the deterministic, discrete version of this problem, as presented in Tanner and Sutton (2005).

We next experimented with predicting a sinusoid, where the observation emitted at time $t$ was given by $o_t = (\sin(0.5t) + 1)/2$. Observations were again corrupted by mean zero Gaussian noise with 0.05 standard deviation. We again let $n = 4$, and $\sigma = 0.3$, but rather than pick a specific chain depth $d$ of the question network for which to present results, we plot the learning curves for a few values of $d$. Each curve is again an average of 30 runs. Figure 6 shows these curves, and it is clear that while increasing the depth of the question network chains up from 1 through 5 improves the quality of the model learned, having depth greater than 5 does not produce very significant performance benefits aside from some slightly faster convergence.

### 5.2 Controlled Systems

We next evaluated our approach on two controlled versions of the dynamical systems used above. In each case, we introduced an action dimension that varied the amplitude of the corresponding wave function. The possible actions for each system were in $[0, 1]$, and resulted in modulating the amplitude from 0 to 1 continuously. That is, for a given action $a \in [0, 1]$, the square wave alternated between emitting values $a + \frac{1-a}{2}$ and $\frac{1-a}{2}$, each five steps at a time. Similarly, the sine wave emitted an observation at time $t$ according to $o_t = \frac{a}{2}(\sin(0.5t) + 1) + \frac{1-a}{2}$, given action $a \in [0, 1]$. Observations in both systems were again corrupted by mean-zero Gaussian noise with standard deviation 0.05.

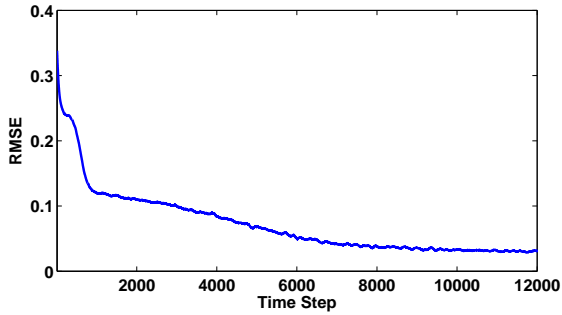

Figure 7: RMSE of the one-step predictions of all $\phi_i \in \Phi$ as a function of time for the noisy, controlled square wave. Each point is an average of the error over the previous 100 time steps.

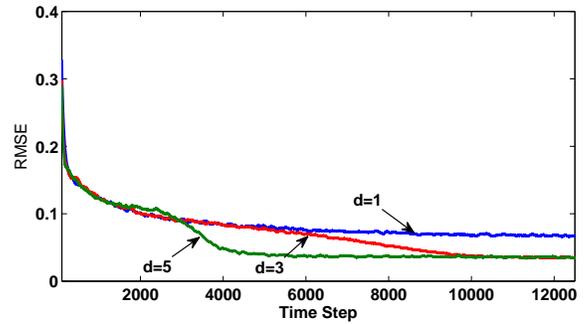

Figure 8: RMSE of the one-step predictions of all $\phi_i \in \Phi$ as a function of time for the noisy, controlled sine wave. Each point is an average of the error over the previous 100 time steps.

In both experiments we let $n = m = 4$, but we found it necessary to use different values of $\sigma$ for the feature functions than for the action activation functions. We set the former, $\sigma_\phi$, to 0.3, and the latter, $\sigma_\psi$, to 0.1. We again set the depth $d$ of the question network chains to 5 in the square wave experiment, and varied the depth of the networks in the sine wave experiment, plotting the results for each depth.

The policies used to collect data were smoothed versions of a random walk over the action space. Errors were computed as above, where the expected values of a given feature function $\phi_i$ were calculated by weighting the predictions of each of the children of $\phi_i$ by the activation of the child's associated action activation function, given the last action taken.

Figures 7 and 8 show the RMSE of one-step predictions for the controlled square wave and sine wave experiments, respectively. We see in Figure 7 that the system is able to learn almost as good a model of the controlled system as it did of the uncontrolled system, indicating that our mechanism for handling continuous actions seems to be viable. Similarly, Figure 8 shows that, although still dependent on having enough steps of prediction, the network is able to learn a good model of the controlled sine wave system as well.

Lastly, we tested our algorithm on a partially observable version of a dynamical system common in the reinforcement learning literature, the mountain car, in which an underpowered car must be driven up a steep cliff in a valley. Because the car is underpowered it cannot drive directly up the hill, but must reverse up the rear side of the valley to gain enough momentum to make it up the far side. We refer the reader to Sutton and Barto (1998) for the details of the dynamics. When the position and velocity of the car are given as observations, the system is fully observable. We thus eliminated the velocity component of the observation, producing a one-dimensional observation which is not sufficient to maintain state. Additionally, as in our other experiments, we corrupted each observation with mean-zero Gaussian noise with standard deviation 0.05.

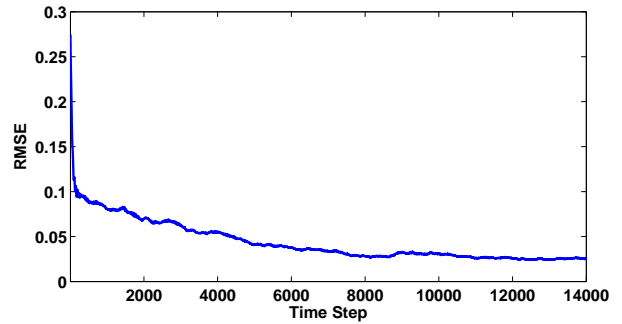

Figure 9: RMSE of the one-step predictions of all $\phi_i \in \Phi$ as a function of time for the noisy, partially observable mountain car task. Each point is an average of the error over the previous 100 time steps.

As in the previous experiments, we let $n = m = 4$ and set $\sigma_\phi$ to 0.3 and $\sigma_\psi$ to 0.1. The chain depth of the question network was set to 5. Figure 9 plots the RMSE of one-step predictions for the mountain car system. We see that the TD network used was able to learn a good model of the dynamics and thus was able to recover the velocity dimension by making use of multi-step predictions.

## 6 Discussion

We have presented an extension to the TD($\lambda$) network learning algorithm that allows one to use a TD network to model partially observable, noisy, continuous dy-namical systems. This represents the first instance of a fully incremental algorithm for learning a predictive representation of a continuous dynamical system. Our results show that the algorithm is capable of learning accurate models of both controlled and uncontrolled versions of such systems that are robust to noise.

In the work presented here we constructed question networks based on intuition and trial-and-error. In general it is desirable to automate this process and discover a (preferably minimal) question network based on a stream of experience. An online discovery algorithm for discrete TD networks was presented in Makino and Tagaki (2008). We chose not to employ it in this work so as to better isolate the contributions of extending TD networks to continuous systems. However, we see no immediate reason why the approach presented there cannot be combined with our extension, though of course this needs to be tested empirically.

Interesting directions for future research include state abstraction in continuous TD networks to facilitate scaling our approach up to higher-dimensional systems. State abstraction in other predictive representation formalisms has been considered for both discrete (Wolfe et al., 2008) and continuous dynamical systems (Wingate, 2008). However, although temporal abstraction in discrete TD networks has been explored recently (Sutton et al., 2006), to our knowledge there has been no work on state abstraction in TD networks. Using features that are defined over every dimension of the observation space is not always necessary for structured environments. Taking advantage of such structure may lead to compact representations that are easier to learn.

Finally, our approach has been agnostic about the observation features used. The accuracy of the model learned will obviously be dependent upon the form those features take. There is a large body of recent work on basis function selection and construction for value function approximation in Markov decision processes (Mahadevan, 2008; Parr et al., 2007), and it may be interesting to consider applying work in those areas to choosing or constructing appropriate features for observation spaces in partially observable, continuous domains.

**Acknowledgements**

The author would like to thank George Konidaris, Kimberly Ferguson, Scott Kuindersma, and Pippin Wolfe for their helpful comments and discussions.

**References**

Littman, M. L., Sutton, R. S., & Singh, S. (2002). Predictive representations of state. *In Advances In Neural Information Processing Systems (NIPS)* (pp. 1555–1561).

Mahadevan, S. (2008). *Representation discovery using harmonic analysis.* Morgan and Claypool.

Makino, T., & Takagi, T. (2008). On-line discovery of temporal-difference networks. *In International Conference on Machine Learning (ICML).* Helsinki, Finland.

Parr, R., Painter-Wakefield, C., Li, L., & Littman, M. (2007). Analyzing feature generation for value-function approximation. *In International Conference on Machine Learning (ICML).*

Rafols, E. J., Ring, M. B., Sutton, R. S., & Tanner, B. (2005). Using predictive representations to improve generalization in reinforcement learning. *In International Joint Conference on Artificial Intelligence (IJCAI)* (pp. 835–840).

Singh, S., & James, M. R. (2004). Predictive state representations: A new theory for modeling dynamical systems. *In Uncertainty in Artificial Intelligence (UAI)* (pp. 512–519).

Sutton, R. S. (1988). Learning to predict by the methods of temporal differences. *Machine Learning* (pp. 9–44).

Sutton, R. S., & Barto, A. G. (1998). *Reinforcement learning: An introduction.* Cambridge, Massachusetts: MIT Press.

Sutton, R. S., Rafols, E. J., & Koop, A. (2006). Temporal abstraction in temporal-difference networks. *In Advances in Neural Information Processing Systems (NIPS)* (pp. 1313–1320).

Sutton, R. S., & Tanner, B. (2005). Temporal-difference networks. *In Advances in Neural Information Processing Systems (NIPS)* (pp. 1377–1384).

Tanner, B., & Sutton, R. S. (2005). Td($\lambda$) networks: temporal-difference networks with eligibility traces. *ICML* (pp. 888–895).

Wingate, D. (2008). *Exponential family predictive representations of state.* Doctoral dissertation, University of Michigan.

Wolfe, B., James, M. R., & Singh, S. (2008). Approximate predictive state representations. *In Autonomous Agents and Multiagent Systems (AAMAS)* (pp. 363–370). Estoril, Portugal.